\documentclass[letterpaper, 10 pt, conference]{ieeeconf}  

\IEEEoverridecommandlockouts                              

\usepackage{amsmath} 
\usepackage{graphicx}
\usepackage{xcolor}
\usepackage{placeins}
\usepackage{todonotes}
\usepackage{booktabs} 
\usepackage{dsfont} 
\usepackage{tabularx} 
\usepackage{subcaption} 
\usepackage{multirow}
\usepackage{hyperref}

\title{\LARGE \bf
When Planners Meet Reality: How Learned, Reactive Traffic Agents Shift nuPlan Benchmarks
}

\author{Steffen Hagedorn$^{1}$, Luka Donkov$^{2}$, Aron Distelzweig$^{3}$ and Alexandru P. Condurache$^{1}$
\thanks{$^{1}$Robert Bosch GmbH,
        Leonberg, Germany and
        Institute for Neuro- and Bioinformatics, University of Lübeck, Germany
        {\tt \small steffen.hagedorn@de.bosch.com}}%
\thanks{$^{2}$Robert Bosch GmbH,
        Feuerbach, Germany and
        Engineering Faculty, DHBW Stuttgart, Germany.}%
\thanks{$^{3}$Department of Computer Science, University of Freiburg, Germany.}%
}

\definecolor{myred}{HTML}{DF7162}
\definecolor{myblue}{HTML}{6A9BDD}

\begin{document}

\maketitle
\thispagestyle{empty}
\pagestyle{empty}

\begin{abstract}
Planner evaluation in closed-loop simulation often uses rule-based traffic agents, whose simplistic and passive be-havior can hide planner deficiencies and bias rankings.
Widely used \texttt{IDM} agents simply follow a lead vehicle and cannot react to vehicles in adjacent lanes, hindering tests of complex interaction capabilities.
We address this issue by integrating the state-of-the-art learned traffic agent model \texttt{SMART} into nuPlan.
Thus, we are the first to evaluate planners under more realistic conditions and quantify how conclusions shift when narrowing the sim-to-real gap.
Our analysis covers 14 recent planners and established baselines and shows that \texttt{IDM}-based simulation overestimates planning performance: nearly all scores deteriorate.
In contrast, many planners interact better than previously assumed and even improve in multi-lane, interaction-heavy scenarios like lane changes or turns.
Methods trained in closed-loop demonstrate the best and most stable driving performance.
However, when reaching their limits in augmented edge-case scenarios, all learned planners degrade abruptly, whereas rule-based planners maintain reasonable basic behavior.
Based on our results, we suggest \texttt{SMART}-reactive simulation as a new standard closed-loop benchmark in nuPlan and release the \texttt{SMART} agents as a drop-in alternative to \texttt{IDM} at \color{cyan}\url{https://github.com/shgd95/InteractiveClosedLoop}.
\end{abstract}

\section{INTRODUCTION}
\FloatBarrier  

\begin{figure}[tp]
    \centering
    \includegraphics[width=\columnwidth]{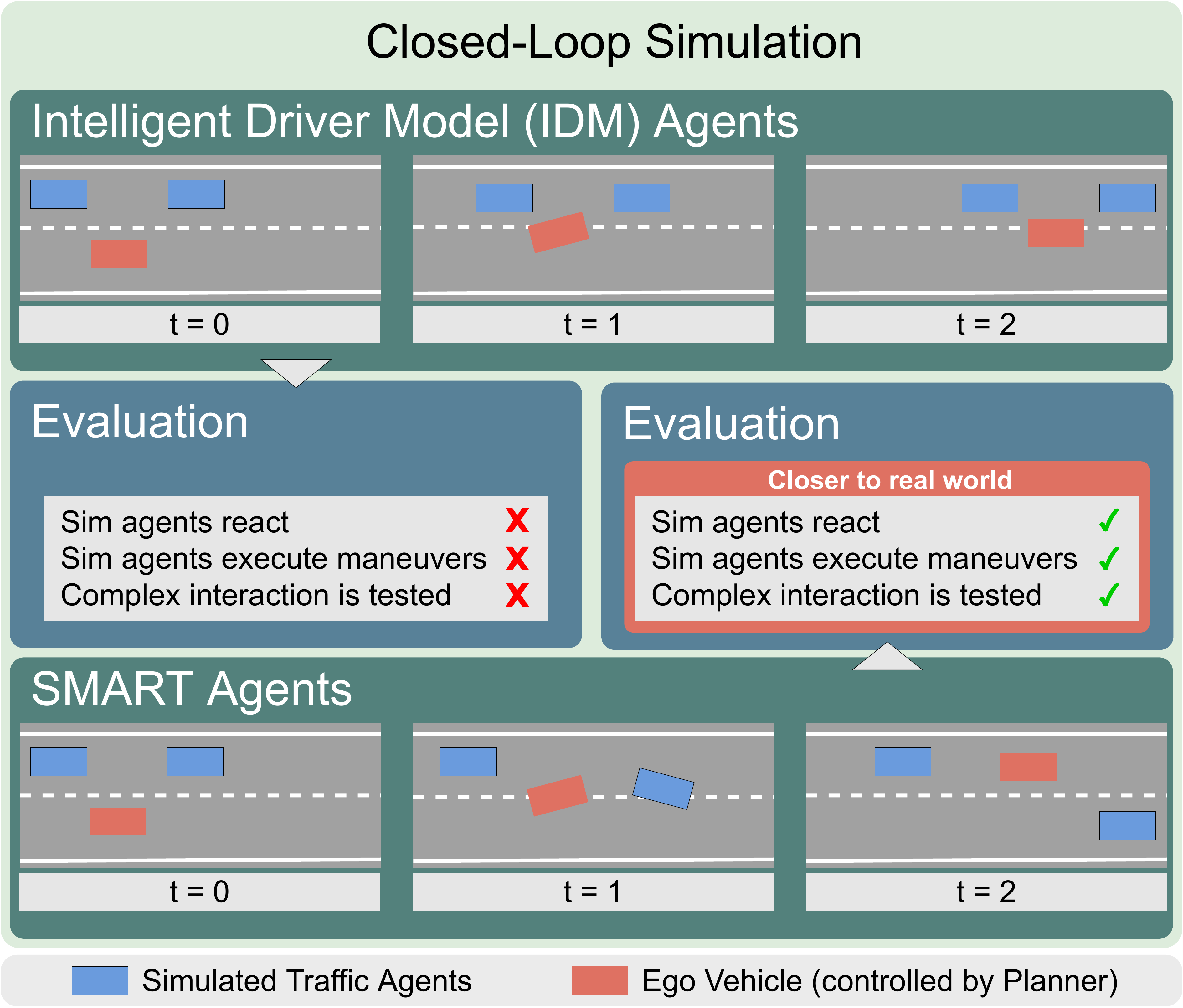}
    \caption{Intelligent Driver Model (\texttt{IDM}) vs. \texttt{SMART} agents: same scene, different world. Starting with the same real-world scene state, the two traffic agent models lead to different scene developments when evaluating the same ego trajectory planner in a closed-loop simulation. Since the \texttt{IDM} cannot see vehicles in adjacent lanes it does not react to the ego vehicle's lane change attempt and behaves passively. In contrast, \texttt{SMART} agents trained from real traffic data behave less passive, react to other agents across lanes and execute driving maneuvers themselves. These characteristics of the sim agents are important for testing complex interactions and for assessing a planner’s driving behavior in real-world traffic.}
    \vspace{-0.5cm}
    \label{fig:cover_figure}
\end{figure}

In the development of automated driving systems it is crucial to assess performance under real-world conditions correctly before commencing tests in real traffic.
The assessment's quality strongly depends on the evaluation procedure.
Open-loop evaluation focuses on imitating a human expert driver as closely as possible. Various studies show that open-loop evaluation does not correlate with real-world driving performance~\cite{codevilla2018offline, dauner2023parting}.
In contrast, closed-loop evaluation provides more generalizable results by simulating the environment's response to actions of the ego vehicle.
Instead of imitating an expert driver, good driving behavior in general is rewarded in terms of safety, progress, and comfort metrics.
Closed-loop simulation produces scene developments that deviate from the ground truth, for example the planner could decide to stay on a lane, whereas the expert driver performed a lane change in that situation, and achieve an equally good score.
However, the generalizability of closed-loop evaluation strongly depends on the quality of the traffic agent model that updates the simulation, in which the planner acts.
The widely used nuPlan framework and its benchmarks~\cite{karnchanachari2024towards} rely on simplistic, rule-based \texttt{IDM} (Intelligent Driver Model)~\cite{treiber2000congested} traffic agents, whose behavior often lacks realism.
For example, \texttt{IDM} agents do not perceive vehicles in adjacent lanes and ignore lane-change attempts (Fig{.}~\ref{fig:cover_figure}) or brake too hard when another car merges anyway.
Such behavior is problematic for various reasons: Planners can aggressively take advantage of this passivity to exploit benchmarks~\cite{hagedorn2024integration}.
Further, evaluation should rather cover the harder case in which other vehicles behave actively and human-like, instead of cautiously and passively.
Finally, simplistic and unimodal traffic agent models prevent complex interactions and create a sim-to-real gap that biases planner benchmarks.
We address this gap by integrating a learned, reactive state-of-the-art traffic agent model into the nuPlan framework to establish a new benchmark based on more realistic closed-loop simulation.
We chose \texttt{SMART}~\cite{wu2024smart} for its strong performance in the Waymo 2024 Sim Agents Challenge with high realism and interaction scores, map compliance, and real-time inference, making it practical for large nuPlan rollouts.

Since changing the traffic agent model fundamentally alters closed-loop simulations, we perform a series of novel experiments to investigate how planners behave under more realistic conditions:
Using the \texttt{SMART} agents we are the first ones to evaluate 14 state-of-the-art planners and common baselines of the nuPlan framework in a learned, reactive traffic simulation.
We compare planner performance in \texttt{SMART}-based and \texttt{IDM}-based closed-loop simulations on three benchmarks:
the comprehensive \textit{Val14}~\cite{dauner2023parting}, the more difficult \textit{Test14-hard}~\cite{cheng2024rethinking}, and the augmented edge-case scenarios of interPlan~\cite{hallgarten2024can}.
We find that \texttt{IDM}-based simulation indeed distorts planner rankings and systematically overestimates planner performance but underestimates interaction capabilities.
Specifically, imitation-learned planners deteriorate in simple scenarios, while rule-based planners deteriorate in harder scenarios that require advanced interactions.
Planners trained in closed-loop also perform better and more stable in realistic closed-loop simulations.
When stress-testing planners in the hardest scenarios of interPlan, all methods reach their limits but rule-based methods degrade smoothly whereas learned planners exhibit a sudden tipping point.

In summary, the main contributions of our work are:
\begin{enumerate}
    \item We are the first ones to evaluate established nuPlan planners in realistic, interactive traffic, revealing a general overestimation of planner performance under \texttt{IDM}-based simulation on diverse benchmarks alongside a simultaneous underestimation of interaction capabilities and further discrepancies.
    \item We propose a new \texttt{SMART}-reactive benchmark for nuPlan to enable the realistic analysis of planner strengths and failure cases and introduce a corresponding closed-loop score.
    \item We provide a community-ready implementation of state-of-the-art learned \texttt{SMART} traffic agents for nuPlan to enable model training and evaluation in realistic, reactive closed-loop simulations.
\end{enumerate}

\section{RELATED WORK}
\subsection{Automated Driving Evaluation}
As it is unsafe and unethical to directly evaluate automated driving systems in public traffic, offline evaluation paradigms have emerged.
Early planner evaluation has largely been \emph{open-loop}: models predict the future based on fixed human-driven logs and are scored by imitation metrics with no feedback from the ego’s actions to the scene.
This setup scales well but suffers from covariate shift and does not reflect real driving quality~\cite{ross2011reduction,codevilla2018offline,dauner2023parting}.

\begin{figure}[tp]
    \centering
    \includegraphics[width=\columnwidth]{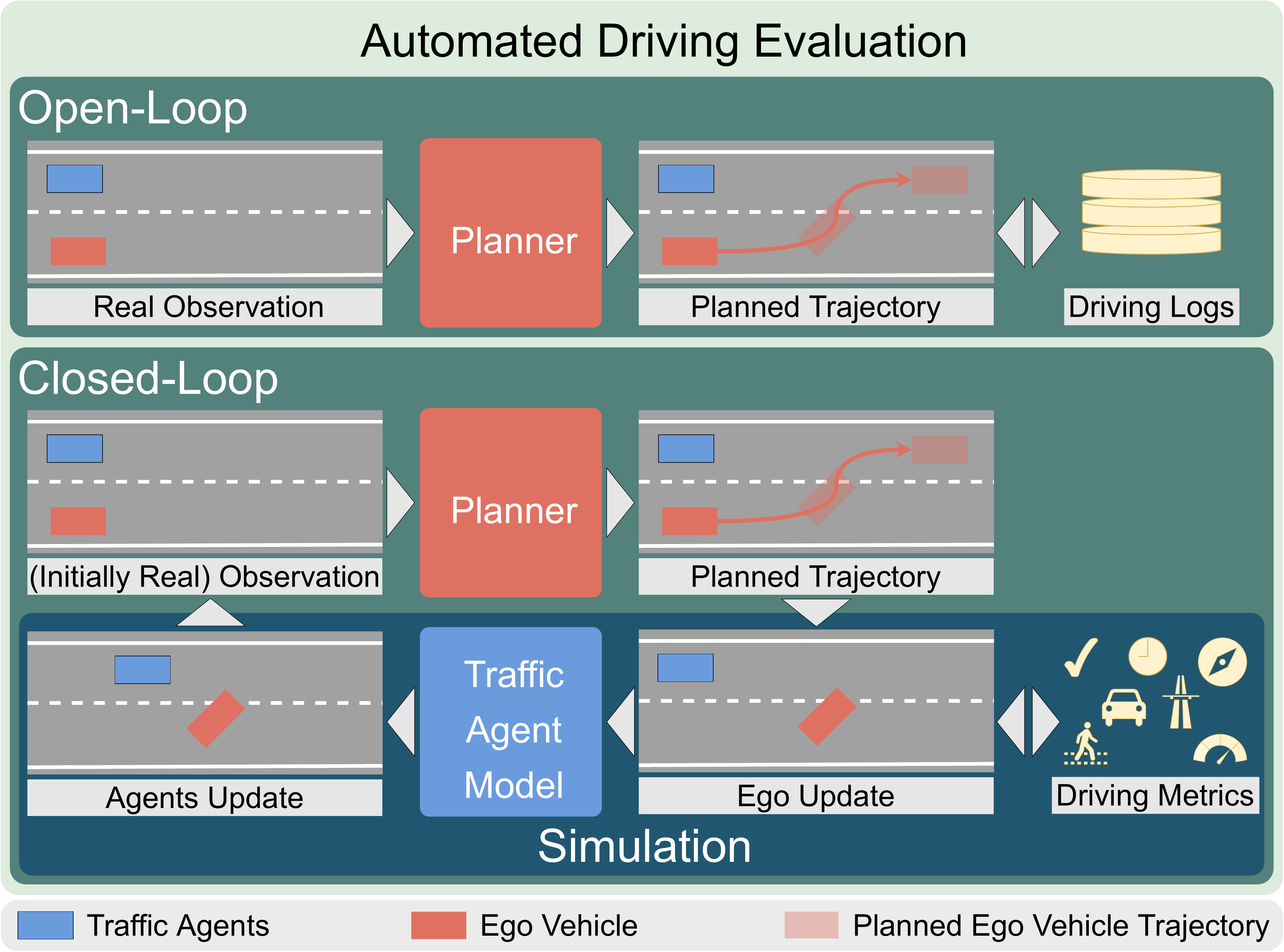}
    \caption{Automated Driving Evaluation. In open-loop evaluation the planner generates a future trajectory from a real observation, which is then compared to an expert's driving log. However, open-loop performance does not correlate with real-world driving performance as it lacks evidence for the system's behavior when deviating from human expert states. Closed-loop evaluation addresses this shortcoming by allowing the scene state to deviate from the recorded real-world data and rewards good driving behavior instead of imitation. Nevertheless, the quality of the traffic agent model determines how well closed-loop performance mirrors real-world driving.}
    \vspace{-0.5cm}
    \label{fig:related_work}
\end{figure}

In contrast, \emph{closed-loop} evaluation restores interaction: the planner controls the ego in simulation and is judged by task success, safety, and comfort as the world reacts to its decisions (Fig{.}~\ref{fig:related_work})~\cite{bouzidi2025closing}.
Within closed-loop, benchmarks distinguish \emph{non-reactive} backgrounds, which replay logged trajectories even if the ego deviates, from \emph{reactive} backgrounds, where other agents respond to the ego.
nuPlan supports both and, by default, realizes reactivity with the rule-based Intelligent Driver Model (\texttt{IDM}) for background traffic~\cite{karnchanachari2024towards,treiber2000congested}.
While simple and interpretable, such rule-based traffic can miss lateral negotiation, multimodality, or irrational (human-like) behavior, and thereby yield optimistic safety and overly stable planner rankings—precisely the gap we study by replacing \texttt{IDM} with learned, reactive agents (\texttt{SMART}) in nuPlan~\cite{wu2024smart}.

Multiple works show that conclusions shift when moving from non-reactive to reactive traffic: planner rankings, failure modes, and safety–efficiency trade-offs can flip~\cite{dauner2023parting,fan2024risk}.
Swapping one reactive background for another also alters scene progress (Fig{.}~\ref{fig:cover_figure}), making the choice of the traffic agent simulation model crucial for the generalizability of evaluation results.

\subsection{Traffic Agent Simulation}
Closed-loop evaluation relies on a traffic simulator to model the scene progress.
Traffic simulators can be classified as sensor-level or object-level methods.
Sensor-level simulators render raw camera, lidar, or radar outputs, enabling full-stack and perception testing but with heavy compute costs.
CARLA, LGSVL, and Bench2Drive are widely used simulators of this category~\cite{dosovitskiy2017carla, rong2020lgsvl, jia2024bench2drive}.
Often sensor-level simulators use an underlying object-level simulation based on which the sensor outputs are rendered.

Object-level simulators update the world state directly without rendering sensor outputs.
They are fast and scale to many scenarios, making them well-suited for planning and interaction studies.
Their speed also enables rapid closed-loop training iterations.
Perception robustness and full-stack performance, however, cannot be tested with object-level simulators.
nuPlan, Nocturne, and WOSAC emphasize this paradigm with multi-agent, interaction-centric metrics and leaderboards~\cite{karnchanachari2024towards,vinitsky2022nocturne,montali2023waymo}.
Object-level simulators can be categorized into:

\emph{Rule-based.}
Classical simulation updates each vehicle individually with hand-crafted rules: \texttt{IDM} computes longitudinal acceleration for car-following and \texttt{MOBIL} decides when to change lanes~\cite{treiber2000congested,kesting2007general}.
In practice, these equations are paired with auxiliary rules for right-of-way, signals, and hand-tuned heuristics for gap acceptance and courtesy.
The approach is simple, interpretable, and works well within a fixed domain.
However, they struggle with lateral negotiation, multi-modal intent, and uncooperative behavior.
They often react late and too strongly to cut-ins and need case-specific logic for traffic rules, signals, and every single maneuver, making broad generalization infeasible and limiting their realism.

\emph{Hybrid.}
Hybrid models combine hand-crafted rules or heuristics with learned components.
For example, \texttt{cogniBOT} adds learned components to a cognitive, rule-informed policy to improve realism while keeping control over simulation parameters~\cite{brostek2024achieving}.
Hybrid approaches aim to guarantee rule-compliance while gaining data-driven fidelity.

\emph{Learned.}
Earlier learned simulators such as \texttt{TrafficSim} showed that multi-agent traffic can be learned from logs and rolled out stably in closed loop~\cite{suo2021trafficsim}.
This line of work paved the way for today’s models.
Recent research shows that training traffic agent models with reinforcement learning in closed-loop rollouts can improve purely imitation-based approaches~\cite{bitzer2024analyzing}.
Nevertheless, the recent trend focuses on imitation learning from logs: alongside generative approaches based on diffusion (\texttt{VBD}~\cite{huang2024versatile}) and variational autoencoders (\texttt{TrafficBots}~\cite{zhang2024trafficbots}), the majority of models uses autoregressive Transformers (\texttt{Trajeglish}~\cite{philion2023trajeglish}, \texttt{MVTE}~\cite{wang2023multiverse}, \texttt{SMART}~\cite{wu2024smart}), \texttt{GUMP}~\cite{hu2024solving}, \texttt{BehaviorGPT}~\cite{zhou2024behaviorgpt}).

\textit{Why we chose \texttt{SMART}.}
We integrate \texttt{SMART} as a new drop-in background for nuPlan simulations. \texttt{SMART} discretizes the vectorized scene and trajectories into spatio-temporal tokens and trains a decoder-only Transformer to predict the next tokens~\cite{wu2024smart}. This design makes inference fast and memory-efficient, which is important for large \texttt{nuPlan} rollouts. \texttt{SMART} also achieved strong realism and interaction scores in the Waymo Sim Agents challenge, making it a good candidate to replace \texttt{IDM} when the goal is realistic evaluation of planners in reactive closed-loop simulation.

\subsection{Trajectory Planning}
While in traffic simulation the goal is to generate a realistic traffic flow, i.e., safe, traffic rule-compliant, and kinematically feasible diverse trajectories, the additional objective of trajectory planning is to move toward a navigation goal.
Planner models span three broad styles, closely related to the design of the overall automated driving system in which they are applied~\cite{hagedorn2024integration}.

\textit{End-to-end} systems map sensor outputs like camera images or lidar point clouds directly to trajectories or controls.
Sensor rendering is required to evaluate these models in closed-loop.
For our experiments in the object-level simulation of nuPlan we exclude end-to-end systems and, instead, compare methods that are natively compatible with nuPlan’s closed-loop simulation~\cite{karnchanachari2024towards}.

\textit{Modular} systems keep a clear split between individual tasks, facilitating evaluation in object-level simulators ~\cite{hagedorn2025learning, lowens2025pseudomaptrainer, hagedorn2024pioneering}.
They can be rule-based or learned.
\textit{Rule-based} approaches specify hand-crafted policies for car-following, lane-change, and search or optimization routines~\cite{treiber2000congested, kesting2007general, dauner2023parting}.
These are often used as a simple baseline for motion planning~\cite{treiber2000congested}.
\textit{Learned} planning modules are optimized on training data instead of relying on hand-crafted rules~\cite{scheel2022urban, renz2022plant, zheng2025diffusion}.
In the regime of learned planners, two training paradigms dominate.
\textit{Imitation learning (IL)} fits policies to human demonstrations and remains the most frequently applied paradigm because of data scale and training stability~\cite{renz2022plant, cheng2024pluto}.
IL can achieve strong closed-loop scores but is sensitive to covariate shift through accumulating errors, leading to states unseen during training without closed-loop feedback~\cite{dauner2023parting, codevilla2018offline}.
To address this issue, \emph{Reinforcement Learning (RL)} trains on closed-loop rollouts to let the model deviate from logged expert states and explore the consequences of its actions~\cite{bitzer2024analyzing}.
Instead of fitting the policy to human demonstrations, RL optimizes a reward term that scores generally desirable driving behavior like collision avoidance, staying on-road, or traffic rule compliance~\cite{jaeger2025carl}.

\textit{Hybrid} planners combine learned components with hand-crafted rules.
One typical combination is to generate an initial plan with a learned model, that is then fine-tuned by explicit optimization algorithms~\cite{chekroun2024mbappe, huang2024dtpp}.
Other methods use rule-based safety layers to enforce drivable area compliance, collision avoidance and traffic rule-adherence~\cite{dauner2023parting}.

For our study we select diverse trajectory planners that cover rule-based and hybrid methods as well as IL-based and RL-based learned modular approaches to investigate how realistic traffic agent simulation affects their performance and if conclusions related to these paradigms can be drawn.

\section{METHODOLOGY}

\subsection{Datasets \& Benchmarks}
We train the \texttt{SMART} model and conduct all planner studies on the nuPlan dataset and benchmark~\cite{karnchanachari2024towards}, which contains $\sim$1{,}300 hours of expert driving with auto-labeled tracks, traffic lights, and scenario tags.
nuPlan’s closed-loop simulator runs each scenario for 15\,s at 10\,Hz in which the ego vehicle is controlled by the candidate planner and a low-level controller.
Background traffic can be run in two modes: \emph{non-reactive} (log replay) and \emph{reactive} (agents controlled by the \texttt{IDM}).
We adopt the \textit{train\_150k} training split that samples 150{,}000 scenarios across all scenario types, and evaluate on the \textit{Val14} benchmark, which includes up to 100 scenarios per scenario type across 14 types, yielding 1{,}090 total scenarios after excluding a small number of reactive runs that fail to initialize.
(This matches the setup used by recent nuPlan papers.)
For additional experiments we use \textit{Test14-hard}, a curated 280-scenario split formed by running 100 candidates per scenario type with the strong rule-based baseline \texttt{PDM-Closed} and selecting the 20 lowest-scoring, i.e., ‘hardest’ cases, emphasizing long-tail interactive failures like in tight merges and unprotected turns~\cite{cheng2024rethinking}.
To stress-test planners under even harder conditions we also perform experiments on the manually augmented \textit{interPlan} benchmark~\cite{hallgarten2024can}. Specifically, we use 30 lane change scenarios of which ten are in low, medium and high-density traffic, each.

To assess the realism of \texttt{SMART} simulation agents beyond nuPlan, we additionally curate results of the the Waymo Open Sim Agents Challenge (WOSAC) 2024, that is based on the Waymo Open Motion Dataset (WOMD)~\cite{montali2023waymo}.
WOMD provides large-scale, object-level trajectories with HD maps, comprising over 100k 20\,s segments at 10\,Hz.

\subsection{Metrics}
We report metrics for two aspects of our study: (i) planner performance and (ii) the realism of simulated traffic agents.

\textit{(i) Planner evaluation.}
nuPlan employs three principal metrics: the \emph{open-loop score (OLS)}, the \emph{non-reactive closed-loop score (CLS-NR)}, and the \emph{reactive closed-loop score (CLS-R)}~\cite{karnchanachari2024towards}.
Consistent with prior findings that open-loop prediction quality weakly correlates with closed-loop driving effectiveness, we focus on closed-loop performance and provide the OLS only for reference.
Closed-loop scores are computed per scenario as a weighted combination of soft metrics: \emph{progress} along the route, \emph{time-to-collision} within bounds, and \emph{comfort} (e.g., lateral/longitudinal jerk), which are subject to hard multipliers that zero the score if \emph{at-fault collisions} or \emph{drivable-area violations} occur~\cite{daunersupplementary}.
Scenario scores are then averaged across the benchmark into a normalized $[0,100]$ composite used for ranking~\cite{karnchanachari2024towards,dauner2023parting}.

\textit{\texttt{SMART}-reactive score.}
To narrow the sim-to-real gap of closed-loop planner evaluation, we propose a new benchmark on the \textit{Val14} and \textit{Test14-hard} dataset splits of nuPlan.
Our benchmark uses the scenarios of \textit{Val14} and \textit{Test14-hard} and introduces a novel \textit{\texttt{SMART}-reactive closed-loop score (CLS-SR)}.
The score is computed from the same metrics as CLS-R but utilizes \texttt{SMART} to update the simulated vehicles instead of \texttt{IDM}.
This allows direct comparison of planners across non-reactive, rule-based reactive, and learned reactive simulation.

\textit{(ii) Simulated-traffic realism.}
For \texttt{SMART} and baselines, we report \emph{Average Displacement Error (ADE)} compared to the logged expert driver ground truth and the \emph{Realism Meta-Metric (RMM)} as defined by WOSAC~\cite{montali2023waymo}.
The RMM rewards realistic traffic flow instead of imitation by aggregating kinematic-based, map-compliance, and interaction-oriented components into a single score~\cite{montali2023waymo}.

\subsection{\texttt{SMART} Integration}
\textit{Training.}
We transfer and train \texttt{SMART} in the nuPlan framework while avoiding architectural changes compared to the published model~\cite{wu2024smart}.
Concretely, we use the 2\,Hz action-token vocabulary with 1024 discrete road and motion tokens, each.
We implement the preprocessing of nuPlan agent and map data to match the preprocessing on WOMD.
Following \texttt{SMART}, we introduce noise in the tokenization pipeline: when discretizing trajectories, we perturb the currently matched token by randomly selecting one of the top\mbox{-}6 nearest tokens to the ground-truth in the vocabulary and continue matching from the perturbed state, which exposes the model to distribution shift and mitigates compounding-errors.
The generated input format is directly compatible with \texttt{SMART} and facilitates training without any adaptions of the model itself.
Our models are trained with a 1\,s history and an 8\,s ground-truth future.
Training follows teacher forcing with cross-entropy for next-token prediction.
We trained the small \texttt{SMART} model with 7M learnable parameters on varying amounts of nuPlan and WOMD data to convergence and compared performance on a separate validation set.
We selected checkpoints by token accuracy.
The model used in all subsequent experiments, and released with this paper, was trained on the nuPlan \textit{train\_150k} split for eight epochs.
On a single NVIDIA H200 GPU with a mini-batch size of four, training required four days.

\textit{Inference \& runtime.}
nuPlan advances the simulation at 10\,Hz.
\texttt{SMART} operates on 2\,Hz tokens, so we integrate it in a receding-horizon fashion: every 0.5\,s we re-encode the current scene (including the last 1\,s of history), run a single decoding step to forecast agent movements, and select the maximum-probability token to make the simulation deterministic.
We then upsample the 2\,Hz tokens to 10\,Hz trajectories using \texttt{SMART}’s upsampling procedure and let nuPlan’s tracker execute that motion for the next 0.5\,s.
This avoids rolling out a full 8\,s token sequence at once, keeps computation bounded, and ensures tight feedback between planner actions and background reactions.
The integration is packaged as a drop-in reactive background for nuPlan, so users can select \texttt{SMART} in the simulator configuration exactly like the standard \texttt{IDM} agents.

\subsection{Planner Models}
We benchmark a diverse set of nuPlan-compatible planners that have public code and released checkpoints, so that results are not confounded by re-training variance.
The selection spans rule-based, hybrid, imitation-learning, and reinforcement-learning approaches, including influential earlier baselines and recent state of the art.

\begin{itemize}
    \item \texttt{IDM}~\cite{treiber2000congested} implements rule-based car-following and serves as a conservative planning baseline provided in nuPlan.
    \item \texttt{PDM-Closed}~\cite{dauner2023parting} is a rule-based, centerline-following planner that assesses and optimizes closed-loop metrics.
    \item \texttt{PDM-Hybrid}~\cite{dauner2023parting} combines a learned ego-forecast with a rule-based refinement stage to improve stability.
    \item \texttt{GameFormer}~\cite{huang2023gameformer} combines a Transformer with game theory to model strategic interactions among traffic participants, paired with a rule-based refinement.
    \item \texttt{DTPP}~\cite{huang2024dtpp} stands for differentiable tree-structured policy planning that jointly learns prediction and planning costs and searches over a trajectory tree.
    \item \texttt{PLUTO}~\cite{cheng2024pluto} implements modular end-to-end imitation learning with vectorized scene encoding and contrastive training, leading to high closed-loop performance.
    \item \texttt{UrbanDriver}~\cite{scheel2022urban} is an early learned baseline integrated in nuPlan, that realizes a policy-gradient planner trained from demonstrations in object-level closed loop.
    \item \texttt{GC-PGP}~\cite{hallgarten2023prediction} extends na\"{i}ve behavior cloning with goal-conditioning at inference time.
    \item \texttt{PlanCNN}~\cite{renz2022plant} realizes a rasterized BEV planner that scores candidate trajectories with a convolutional backbone and chooses the best.
    \item \texttt{PlanTF}~\cite{cheng2024rethinking} is an early Transformer-based planner operating on vectorized map and agent features.
    \item \texttt{PDM-Open}~\cite{dauner2023parting} is a fully learned, simplistic variant of the \texttt{PDM} conditioned on a reference path and ego state.
    \item \texttt{Diffusion Planner}~\cite{zheng2025diffusion} applies a generative model that samples ego trajectories via diffusion while enforcing map and interaction constraints.
    \item \texttt{CaRL}~\cite{jaeger2025carl} implements a large-scale reinforcement-learning planner optimized directly in closed-loop training with simple rewards.
\end{itemize}

To enable fair evaluation in nuPlan’s closed-loop simulator, we use each author’s recommended checkpoint and default evaluation configuration.
Where a model offers multiple variants, we evaluate the authors’ recommended/published variant.

\begin{figure*}[tp]
    \centering
    \includegraphics[width=\textwidth]{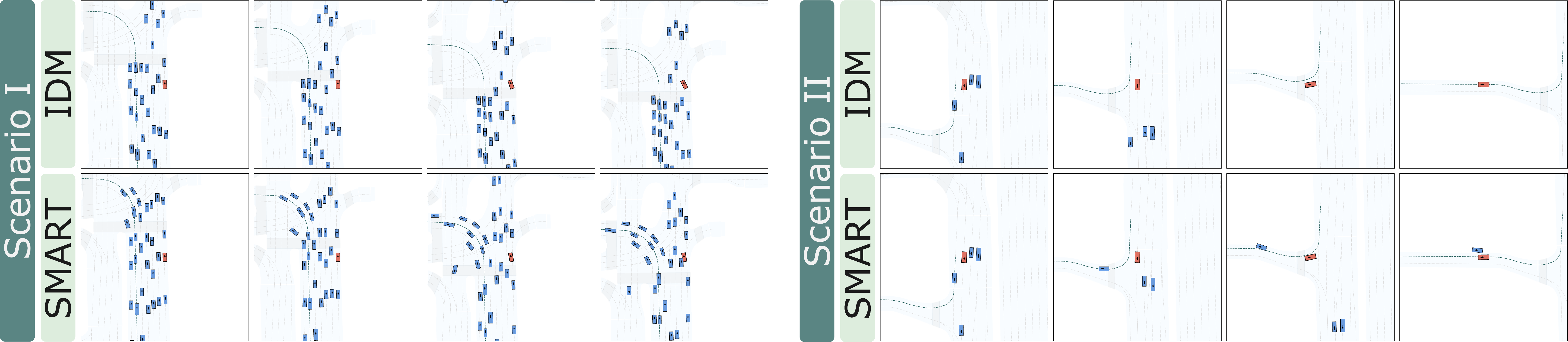}
    \caption{Closed-loop simulation with \texttt{SMART} agents compared to \texttt{IDM} agents in two exemplary scenarios of interPlan-lane-change. In scenario I the ego (red) must execute multiple lane changes in dense traffic to follow the intended route (dashed). \texttt{SMART} agents show more diverse behavior and decreased passivity compared to \texttt{IDM}. The ego surprises the other vehicle with its sudden lane change inside the intersection, causing a collision. In scenario II a \texttt{SMART} agent turns right and then stops at the side of the road. The ego vehicle nudges around it after turning, proving its interaction capability, which remains untested when using \texttt{IDM} agents. In all four simulations the ego vehicle is steered by the best performing planner \texttt{CaRL}.}
    \label{fig:qualitative_comparison}
\end{figure*}

\section{EXPERIMENTS \& RESULTS}
Running the first ever closed-loop planner evaluation with realistically reactive traffic agents on nuPlan yields the results shown in Table~\ref{tab:planner_reeval}.
The comparison comprises 14 diverse planners and reports our novel \texttt{SMART}-reactive closed-loop score (CLS-SR), alongside the standard \texttt{IDM}-reactive score (CLS-R) and the non-reactive log-replay score (CLS-NR).
Open-loop scores are provided for reference; the focus of our analysis is on closed-loop simulations.
All closed-loop runs were executed by us under identical simulator settings to ensure comparability.

We structure the analysis around key insights emerging from these results.

\begin{table*}[tp]
    \centering
    \resizebox{1.0\textwidth}{!}{
    \begin{tabular}{lllrrrrrr}
        \toprule
        \multicolumn{3}{c}{\textbf{Planner}} & \multicolumn{4}{c}{\textbf{Val14}} & \multicolumn{2}{c}{\textbf{Test14-hard}} \\
        \cmidrule(lr){1-3}\cmidrule(lr){4-7}\cmidrule(lr){8-9}
        \textbf{Type} & \textbf{Paradigm} & \textbf{Method} & \textbf{OLS} $\uparrow$ & \textbf{CLS-NR} $\uparrow$ & \textbf{CLS-R} $\uparrow$ & \textbf{CLS-SR} $\uparrow$ & \textbf{CLS-R} $\uparrow$ & \textbf{CLS-SR} $\uparrow$ \\
        \midrule
        \textcolor{gray}{Expert Log} & \textcolor{gray}{Human} & \textcolor{gray}{\texttt{Log Replay}~\cite{karnchanachari2024towards}} & \textcolor{gray}{100} & \textcolor{gray}{94} & \textcolor{gray}{80} & \textcolor{gray}{76} (\textcolor{myred}{-4}) & \textcolor{gray}{69} & \textcolor{gray}{64} (\textcolor{myred}{-5}) \\
        \midrule
        Rule-Based & Rule & \texttt{IDM Planner}~\cite{treiber2000congested} & 38 & 76 & 77 & 77 (0) & 62 & 55 (\textcolor{myred}{-7}) \\
        Rule-Based & Rule & \texttt{PDM-Closed}~\cite{dauner2023parting}  & 42 & 93 & 93 & 89 (\textcolor{myred}{-4}) & 76 & 74 (\textcolor{myred}{-2}) \\
        \midrule
        Hybrid & IL + Rule & \texttt{PDM-Hybrid}~\cite{dauner2023parting}  & 84 & 93 & 93 & 89 (\textcolor{myred}{-4}) & 76 & 72 (\textcolor{myred}{-4}) \\
        Hybrid & IL + Rule & \texttt{GameFormer}~\cite{huang2023gameformer}  & 82 & 81 & 82 & 78 (\textcolor{myred}{-4}) & 70 & 62 (\textcolor{myred}{-8}) \\
        Hybrid & IL + Rule & \texttt{DTPP}~\cite{huang2024dtpp}  & 61 & 64 & 64 & 62 (\textcolor{myred}{-2}) & 47 & 47 (0) \\
        Hybrid & IL + Rule & \texttt{PLUTO}~\cite{cheng2024pluto}  & --- & 93 & 87 & 84 (\textcolor{myred}{-3}) & 76 & 69 (\textcolor{myred}{-7}) \\
        \midrule
        Learned & IL & \texttt{Urban Driver}~\cite{scheel2022urban} & 82 & 53 & 51 & 43 (\textcolor{myred}{-8}) & 41 & 38 (\textcolor{myred}{-3}) \\
        Learned & IL & \texttt{GC-PGP}~\cite{hallgarten2023prediction} & 83 & 59 & 56 & 50 (\textcolor{myred}{-6}) & 43 & 41 (\textcolor{myred}{-2}) \\
        Learned & IL & \texttt{PlanCNN}~\cite{renz2022plant} & 64 & 73 & 70 & 65 (\textcolor{myred}{-5}) & 58 & 51 (\textcolor{myred}{-7}) \\
        Learned & IL & \texttt{PlanTF}~\cite{cheng2024rethinking} & 89 & 85 & 77 & 72 (\textcolor{myred}{-5}) & 59 & 58 (\textcolor{myred}{-1}) \\
        Learned & IL & \texttt{PDM-Open}~\cite{dauner2023parting} & 86 & 50 & 55 & 53 (\textcolor{myred}{-2}) & 37 & 38 (\textcolor{myblue}{+1}) \\
        Learned & IL & \texttt{Diffusion Planner}~\cite{zheng2025diffusion}  & --- & 90 & 83 & 78 (\textcolor{myred}{-5}) & 68 & 63 (\textcolor{myred}{-5}) \\
        \midrule
        Learned & RL & \texttt{CaRL}~\cite{jaeger2025carl}  & --- & 94 & 93 & 90 (\textcolor{myred}{-3}) & 85 & 82 (\textcolor{myred}{-3}) \\
        \bottomrule
    \end{tabular}
    }
    \vspace{0.2cm}
    \caption{\textbf{Planner evaluation on nuPlan benchmarks.} Open-loop (OLS) and closed-loop non-reactive (CLS-NR) results are reported according to~\cite{dauner2023parting, renz2022plant}, and~\cite{zheng2025diffusion}, while all closed-loop reactive (CLS-R) and closed-loop \texttt{SMART} reactive (CLS-SR) scores are from our experiments. CLS-SR is measured using our \texttt{SMART} sim agents.}
    \vspace{-0.5cm}
    \label{tab:planner_reeval}
\end{table*}

\begin{table}[t]
\centering
\resizebox{1.0\columnwidth}{!}{
\begin{tabular}{lcc cc cc}
\toprule
& \multicolumn{6}{c}{\textbf{CLS} $\uparrow$}
\\
\cmidrule(lr){2-7}
& \multicolumn{2}{c}{\textbf{Low Density}} & \multicolumn{2}{c}{\textbf{Mid Density}} & \multicolumn{2}{c}{\textbf{High Density}} \\
\cmidrule(lr){2-3}\cmidrule(lr){4-5}\cmidrule(lr){6-7}
\textbf{Planner} & \texttt{IDM} & \texttt{SMART} & \texttt{IDM} & \texttt{SMART} & \texttt{IDM} & \texttt{SMART} \\
\midrule
\texttt{IDM}          & 63 & 59 (\textcolor{myred}{-4})   & 62 & 58 (\textcolor{myred}{-4})   & 63 & 53 (\textcolor{myred}{-10}) \\
\texttt{PDM-Closed}   & 61 & 62 (\textcolor{myblue}{+1}) & 62 & 54 (\textcolor{myred}{-8})   & 62 & 46 (\textcolor{myred}{-16}) \\
\midrule
\texttt{PDM-Hybrid}   & 62 & 62 (0)      & 62 & 55 (\textcolor{myred}{-7})   & 61 & 46 (\textcolor{myred}{-15}) \\
\texttt{GameFormer}   & 16 & 30 (\textcolor{myblue}{+14}) & 47 & 47 (0) & 28 & 37 (\textcolor{myblue}{+9}) \\
\texttt{DTPP}         & 65 & 58 (\textcolor{myred}{-7})   & 66 & 56 (\textcolor{myred}{-10})  & 73 & 24 (\textcolor{myred}{-49}) \\
\texttt{PLUTO}        & 67 & 66 (\textcolor{myred}{-1})   & 42 & 43 (\textcolor{myblue}{+1}) & 48 & 37 (\textcolor{myred}{-11}) \\
\midrule
\texttt{Urban Driver} & 0  & 0 (0)       & 29 & 15 (\textcolor{myred}{-14})  & 0  & 0 (0) \\
\texttt{GC-PGP}       & 27 & 35 (\textcolor{myblue}{+8}) & 17 & 10 (\textcolor{myred}{-7})   & 61 & 0 (\textcolor{myred}{-61})  \\
\texttt{PlanCNN}      & 46 & 56 (\textcolor{myblue}{+10}) & 68 & 12 (\textcolor{myred}{-56})  & 61 & 0 (\textcolor{myred}{-61})  \\
\texttt{PlanTF}       & 50 & 34 (\textcolor{myred}{-16})  & 41 & 37 (\textcolor{myred}{-4})   & 74 & 42 (\textcolor{myred}{-32}) \\
\texttt{PDM-Open}     & 23 & 33 (\textcolor{myblue}{+10}) & 21 & 14 (\textcolor{myred}{-7})   & 26 & 0 (\textcolor{myred}{-26})  \\
\texttt{Diff. Planner}  & 40 & 21 (\textcolor{myred}{-19})  & 21 & 40 (\textcolor{myblue}{+19}) & 16 & 27 (\textcolor{myblue}{+11}) \\
\midrule
\texttt{CaRL}         & 63 & 69 (\textcolor{myblue}{+6})  & 38 & 49 (\textcolor{myblue}{+11}) & 66 & 29 (\textcolor{myred}{-37}) \\
\bottomrule
\end{tabular}
}
\caption{\textbf{Closed-loop scores on the interPlan~\cite{hallgarten2024can} lane change benchmark.} The values reported for \texttt{SMART} agents correspond to CLS-SR and those for \texttt{IDM} to CLS-R. Rule-based planners degrade smoothly while most methods with learned components hit sudden tipping points when inferred too far outside of the training distribution (mid/high density).}
\label{tab:interplan_results}
\end{table}

\textbf{Imitation-learned planners deteriorate on simple scenarios, rule-based planners on hard ones.}
Table~\ref{tab:planner_reeval} shows that, on the standard \textit{Val14} benchmark, imitation-learned planners experience the most considerable drop when moving from \texttt{IDM} agents to \texttt{SMART} agents in closed-loop evaluation (difference of CLS\mbox{-}SR $-$ CLS\mbox{-}R).
On average, imitation-learned planners have a $-5.17$ decreased closed-loop score, whereas hybrid models only decrease by $-3.25$ and rule-based planners by just $-2.0$.
This highlights the advantage of injecting rule-based knowledge to guarantee collision avoidance and drivable-area compliance and to bound comfort-related quantities such as jerk.
A notable exception among the learned methods is \texttt{PDM-Open}, which remains nearly constant across backgrounds.
We attribute this to its minimalistic inputs (ego state and centerline), which effectively encode a centerline-following prior: while unsuited to maximize the absolute score it yields stable behavior.

\textit{Test14-hard} addresses the imbalance of \textit{Val14} toward easy scenarios by selecting the 20\% lowest-scoring cases per type under a strong baseline (\texttt{PDM-Closed}).
Repeating our experiments on \textit{Test14-hard} reverses the trend: imitation-learned planners lose less performance ($-2.83$) than hybrid ($-4.75$) and rule-based ($-4.5$) methods.
Hard scenarios require nuanced interaction skills like gap negotiation, multi-lane merges, unprotected turns, that fixed rules struggle to express.
Our conclusion is that rule-based structure stabilizes behavior on simple to medium difficulty, but does not generalize as well to challenging interaction-heavy scenes.
Nevertheless, all purely imitation-learned planners are still outperformed by rule-based and hybrids methods when looking at the absolute CLS-SR.

\textbf{Closed-loop training enables stable high-quality performance.}
As outlined above, even the state-of-the-art IL model \texttt{Diffusion Planner} cannot match the performance of the rule-based \texttt{PDM-Closed}.
Our experiments with the RL model \texttt{CaRL} confirm that this finding is related to the covariate shift between open-loop training and closed-loop application of the selected IL models (Table~\ref{tab:planner_reeval}).
\texttt{CaRL} is the only available RL model that is trained on closed-loop rollouts of the nuPlan simulator.
While it already matched the performance of \texttt{PDM-Closed} on Val14-CLS\mbox{-}R it ranks first on our \texttt{SMART}-based \textit{Val14}-CLS\mbox{-}SR benchmark and clearly stands out against all other models on \textit{Test14-hard}-CLS-SR with a +8 margin on the second-placed \texttt{PDM-Closed}.
The performance is also more stable than that of many IL, hybrid, and rule-based planners when switching from \texttt{IDM} to \texttt{SMART}-based simulation.
We relate these results to (i)~the property of reinforcement learning that rule-based knowledge can be injected via reward shaping while still learning a flexible policy from large-scale datasets and especially (ii)~the exposure of the policy to its actions in closed-loop rollouts during training, effectively addressing the covariate-shift problem.
Our experiments show that after years of \texttt{PDM-Closed} leading nuPlan benchmarks, \texttt{CaRL} finally clearly outperforms the rule-based method.

\begin{figure*}[tp]
    \centering
    \includegraphics[page=1, clip, trim=65mm 22mm 0mm 0mm, width=\textwidth]{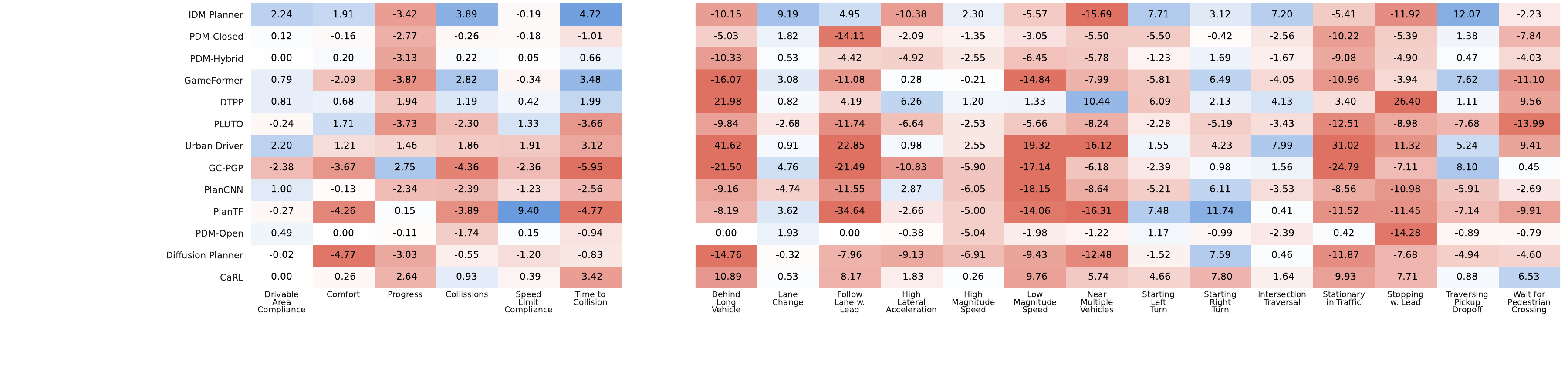}
    \caption{Planner performance shift under \texttt{SMART} agents compared to \texttt{IDM} agents on \textit{Val14} (CLS-SR $-$ CLS-R). The left plot shows a metric-wise breakdown per planner. The right plot provides a scenario-wise analysis.}
    \vspace{-0.25cm}
    \label{fig:planner_details_heatmap}
\end{figure*}

\textbf{\texttt{IDM} agents distort planning benchmarks.}
To understand why performance deteriorates when switching from an \texttt{IDM} background to \texttt{SMART}, we decompose the composite CLS into its components and break down results by scenario type on \textit{Val14} (Fig{.}~\ref{fig:planner_details_heatmap}, left).
Learned planners benefit disproportionately from \texttt{IDM}'s passive and cautious behavior: time-to-collision, ego progress, comfort, and at-fault collisions degrade most under \texttt{SMART}.
This quantifies that inflated planner scores can be facilitated by \texttt{IDM} creating large gaps and yielding early.
In contrast, drivable-area compliance and speed-limit compliance metrics, that are less directly tied to the behavior of other drivers, remain comparatively stable.

The scenario-wise analysis (Fig{.}~\ref{fig:planner_details_heatmap}, right) reveals a consistent pattern.
In multi-lane, interaction-heavy settings like lane change, starting right/left turn, starting straight traffic-light intersection traversal, and traversing pickup/dropoff, scores decrease less or even increase under \texttt{SMART}.
The reason is that \texttt{IDM} primarily reacts to the lead vehicle in-lane, whereas \texttt{SMART} attends to all nearby traffic, enabling realistic cross-lane interactions.
The extreme case occurs when an \texttt{IDM} planner is evaluated against \texttt{IDM} agents: neither side perceives adjacent-lane vehicles, leading to collisions and abrupt driving actions when new vehicles suddenly 'appear' in the lane.
Switching to \texttt{SMART} agents can raise the same planner’s CLS by up to 12\%.
In summary, many planners interact better than their \texttt{IDM}-based scores suggest.
Low scores often rather reflect deficiencies of the \texttt{IDM} background.

\textbf{Beyond their limits learned planners degrade abruptly.}
Finally, we stress-test planners on augmented interPlan lane-change scenarios that vary traffic density (Table~\ref{tab:interplan_results}).
interPlan increases density by adding agents to real-world scenarios while preserving map and route structure.
At low density, lane changes remain feasible, and several planners improve under \texttt{SMART} due to more cooperative across-lane reactions.
As density increases, absolute performance drops sharply for nearly all methods: gaps close, negotiations become contested, and safe merges disappear.
\texttt{SMART} benefits only the most recent learned methods (\texttt{PLUTO}, \texttt{Diffusion Planner}, \texttt{CaRL}) at mid density, but most planners deteriorate as interaction complexity and contention rise.
At high density, lane changes are often infeasible: \texttt{SMART} agents still behave realistically, but the task becomes near-impossible (Fig{.}~\ref{fig:qualitative_comparison}, Scenario~I).
Rule-based and hybrid policies fail more gracefully (e.g., declining to change lanes), whereas fully learned methods degrade more abruptly—consistent with being pushed far out-of-distribution.
Notably, \texttt{CaRL}, that is trained with \texttt{IDM} agents, clearly loses performance at high density when switching to \texttt{SMART} agents, underscoring sensitivity to the training background and the need for diverse, realistic agents during training.
We suggest that using \texttt{SMART} agents not only for evaluation but already in closed-loop training could advance planning further.

\begin{table}[tp]
    \centering
    \resizebox{\columnwidth}{!}{
    \begin{tabular}{lrrrrr}
        \toprule
         & \multicolumn{1}{c}{\textbf{nuPlan Val14}} & \multicolumn{4}{c}{\textbf{WOSAC 2024}} \\
        \cmidrule(lr){2-2}\cmidrule(lr){3-6}
        \textbf{Method} & \textbf{ADE@8s} $\downarrow$ & \textbf{RMM} $\uparrow$ & \textbf{Kinematic} $\uparrow$ & \textbf{Interactive} $\uparrow$ & \textbf{Map} $\uparrow$ \\
        \midrule
        \texttt{IDM}~\cite{treiber2000congested} & 9.60 & 0.62 & 0.48 & 0.72 & 0.56 \\
        \texttt{SMART 7M}~\cite{wu2024smart} & 0.75 & 0.76 & 0.48 & 0.80 & 0.86 \\
        \bottomrule
    \end{tabular}
    }
    \vspace{0.2cm}
    \caption{\textbf{Sim agent performance across nuPlan and WOMD benchmarks.} \texttt{SMART} provides indeed more realistic agents than \texttt{IDM}. nuPlan \texttt{IDM} results are based on~\cite{sun2023large}, WOMD results are curated from~\cite{wu2024smart, xu2025unitsg}.}
    \vspace{-0.5cm}
    \label{tab:sim_agent_performance}
\end{table}

\textbf{\texttt{SMART} agents narrow the sim-to-real gap.}
All experiments rest on the premise that a learned, reactive simulator such as \texttt{SMART} more accurately reproduces real traffic than rule-based \texttt{IDM}.
We therefore compare these two models directly.
Fig{.}~\ref{fig:qualitative_comparison} shows a qualitative comparison of traffic flow when using the two models for traffic agent simulation.
\texttt{SMART} agents demonstrate more versatile behavior than \texttt{IDM} agents, including U-turns, lane changes, and even stopping at the road side.
Especially the \texttt{SMART} agents on the turning lanes are less passive and enter the intersection much sooner (complying with traffic lights).

To quantify these observations, we compiled published metrics where available and reproduced any missing quantities under the same protocol.
First, we consider ADE over 8\,s to evaluate the imitation quality.
As summarized in Table~\ref{tab:sim_agent_performance}, the \texttt{SMART} agents clearly imitate real-world traffic logs more accurately (0{.}75\,m) than the rule-based \texttt{IDM} agents (9{.}6\,m).

However, pure imitation accuracy is an imperfect measure for simulation quality, since future trajectories are inherently multi-modal and many distinct futures can be acceptable.
We therefore complement ADE with the realism meta metric, a realism-oriented composite that evaluates agent behavior along three axes: kinematics, interactions, and map compliance.
Table~\ref{tab:sim_agent_performance} shows that \texttt{SMART} is on par with \texttt{IDM} for kinematic consistency, exhibits stronger interactions between agents, and considerably improves map compliance, reducing off-road events and maintaining larger, human-like distances to road edges.
While \texttt{SMART} is not flawless, these results support its use as a more realistic reactive background than \texttt{IDM}.
Evaluating planners under \texttt{SMART} therefore narrows the sim-to-real gap and enables a more faithful assessment of planning behavior.

\section{CONCLUSION AND FUTURE WORK}
Current nuPlan closed-loop evaluations rely on simple, rule-based reactive traffic (\texttt{IDM}), that can bias results and distort rankings.
We addressed this by integrating learned reactive \texttt{SMART} agents into nuPlan and introducing a more realistic benchmark for nuPlan.
On the new benchmark we evaluated 14 established planners and compared them to their scores in \texttt{IDM} simulation.
Our experiments confirm that \texttt{IDM} shifts benchmarks and is often the origin of planners' poor interaction results, whereas \texttt{SMART} narrows the sim-to-real gap.
We further find that: (i)~imitation-learned planners tend to deteriorate on simple scenarios while rule-based planners deteriorate on hard, interaction-heavy ones; (ii)~closed-loop training yields more stable, high-quality driving; and (iii)~when pushed beyond their limits, learned planners degrade abruptly.
Overall, the reinforcement-learned \texttt{CaRL} performs best in our simulations and clearly surpasses long-time baseline \texttt{PDM-Closed} on \textit{Test14-hard}.
By releasing drop-in \texttt{SMART} agents for nuPlan and proposing standardized reporting (CLS-SR), we enable the community to compare planners under traffic that responds credibly to the ego.
Promising next steps include training \texttt{CaRL} directly against \texttt{SMART} agents and probing robustness with non-deterministic agent behavior, which our implementation supports.






\section*{ACKNOWLEDGMENT}
We thank the authors of \texttt{SMART} for providing their codebase, on which we built our work, and for the helpful exchanges that saved us from running into experimental dead ends.
We also thank our colleague Bin Yang for facilitating our communication with the \texttt{SMART} authors.


\bibliographystyle{IEEEtran}
\bibliography{literature}

\end{document}